\documentclass[journal]{IEEEtran}

\usepackage{amsmath}
\usepackage{amsfonts}
\usepackage{siunitx}
\usepackage{todonotes}
\usepackage[capitalize]{cleveref}
\usepackage{standalone}
\usepackage{booktabs}
\usepackage[caption=false]{subfig}
\usepackage{pgfplots}
\usepackage{enumitem}

\usepackage[natbib,
            bibstyle=ieee,
            doi=false,
            isbn=false
           ]{biblatex}

\AtEveryBibitem{
	\ifentrytype{inproceedings}{
		\clearfield{url}  
		\clearfield{doi}  
	}{}
}

\newcommand{\reals}{\ensuremath{\mathbb{R}}}
\newcommand{\sspace}{\ensuremath{\mathcal{S}}}
\newcommand{\aspace}{\ensuremath{\mathcal{A}}}
\newcommand{\ospace}{\ensuremath{\mathcal{O}}}
\newcommand{\tdist}{\ensuremath{\mathcal{T}}}
\newcommand{\odist}{\ensuremath{\mathcal{Z}}}
\newcommand{\reward}{\ensuremath{\mathcal{R}}}

\newcommand{\beh}{\theta}
\newcommand{\Beh}{\Theta}
\newcommand{\phys}{\ensuremath{q}}
\newcommand{\dt}{\ensuremath{\Delta t}}
\newcommand{\bmax}{\ensuremath{b_\text{max}}}
\newcommand{\ith}[1]{{#1}_i}

\newcommand{\ego}[1]{{#1}_e}
\newcommand{\av}{ego} 
\newcommand{\accelparam}{\bar{a}}

\newcommand{\appropto}{\mathrel{\vcenter{
  \offinterlineskip\halign{\hfil$##$\cr
    \propto\cr\noalign{\kern2pt}\sim\cr\noalign{\kern-2pt}}}}}

\pgfplotsset{every axis legend/.append style={legend cell align=left}}

\begin{document}
\title{Improving Automated Driving through\\POMDP Planning with Human Internal States}

\author{Zachary Sunberg,~\IEEEmembership{Member,~IEEE,}
        Mykel Kochenderfer,~\IEEEmembership{Senior Member,~IEEE}
\thanks{Z. Sunberg is with the Smead Aerospace Engineering Sciences Department at the University of Colorado, Boulder, Colorado, USA; and M. Kochenderfer is with the Department of Aeronautics and Astronautics at Stanford University, Stanford, California, USA.}
}

\maketitle
\IEEEpeerreviewmaketitle

\begin{abstract}
    This work examines the hypothesis that partially observable Markov decision process (POMDP) planning with human driver internal states can significantly improve both safety and efficiency in autonomous freeway driving.
We evaluate this hypothesis in a simulated scenario where an autonomous car must safely perform three lane changes in rapid succession.
Approximate POMDP solutions are obtained through the partially observable Monte Carlo planning with observation widening (POMCPOW) algorithm.
This approach outperforms over-confident and conservative MDP baselines and matches or outperforms QMDP.
Relative to the MDP baselines, POMCPOW typically cuts the rate of unsafe situations in half or increases the success rate by 50\%.%
\end{abstract}

\section{Introduction}


\IEEEPARstart{T}{here} are many criteria that an autonomous vehicle may be judged by, but two of the most important are safety and efficiency.
Unfortunately, safety and efficiency often oppose one another.
Our hypothesis is that planning techniques that consider \emph{internal states} such as intentions and dispositions of other drivers can simultaneously improve safety and efficiency.
An important concept for investigating this hypothesis is the distinction between \emph{epistemic} uncertainty that can be reduced with Bayesian reasoning about observations and \emph{aleatoric} uncertainty that is fundamentally random.
This work evaluates the benefit of modeling internal states with an epistemic uncertainty framework.

The partially observable Markov decision process (POMDP) framework provides a way to formulate a planning problem that considers both epistemic uncertainty in the form of incomplete knowledge of the state and aleatoric uncertainty in the form of noise in individual state transition and observation outcomes~\cite{sabbadin2007purely}.
Many researchers have explored human behavior POMDPs in autonomous driving scenarios (e.g. \cite{hubmann2018automated,sadigh2016leverage,schmerling2017multimodal,bouton2017belief,ulbrich2013probabilistic}).
This work aims to provide insight into the following questions
in the context of a challenging freeway multiple lane change scenario (\cref{fig:pomcpow}):

\begin{figure*}[tb]
    \centering
    \includegraphics[width=\linewidth]{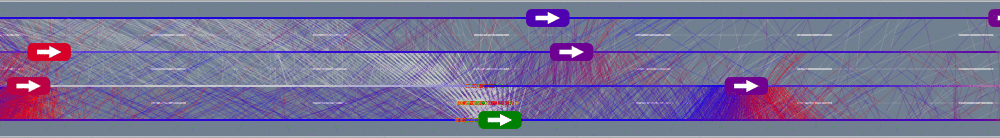}
    \caption{Planning with human internal states using POMCPOW. The green car is the ego vehicle. The color of the other cars indicates the average value of their true internal parameters: redder drivers are more aggressive and bluer drivers more timid. The lines indicate the predicted trajectories of the vehicles in the tree, including rollouts. These trajectories are plotted in a frame moving at the \av's current speed, so trajectories going backwards indicate slower future movement. White lines are the ego vehicle trajectories; colored lines indicate trajectories of other vehicles with the color indicating the sampled parameter values for that vehicle in the tree. Small circles ranging from red to green indicate the estimated value at action nodes on the \av{} trajectories. The absence of action nodes far from the vehicle indicates that the tree only needs a few steps of depth for planning and the majority of each trajectory occurs in rollouts.}
    \label{fig:pomcpow}
    \vspace{-1em}
\end{figure*}

\begin{enumerate}[wide]
    \item \textbf{{Can POMDP planning significantly improve safety and efficiency?}}
        While there are many impressive qualitative results in the literature \cite{sadigh2016leverage,hubmann2018automated,schmerling2017multimodal,ulbrich2013probabilistic}, these studies often do not include systematic quantitative evidence that a POMDP approach can significantly outperform a simpler MDP approach, even in an idealized simulation environment.
        This work isolates the uncertainty model as the experimental variable, comparing state-uncertainty-aware POMDP solution methods to similar MDP methods that consider all uncertainty to be aleatoric.
    \item \textbf{{Can useful POMDP solutions be found in real time?}}
        Several researchers have demonstrated that useful POMDP solutions are readily computable \cite{bai2015intention}, but this work demonstrates this on larger problems with a variable number of vehicles that may be encountered on a multilane highway.
    \item \textbf{{How are the performance advantages of POMDP planning related to the distribution of driver behaviors?}}
        This work shows that correlation between different aspects of driving behavior within individual drivers has a large effect on the relative performance of different planning methods. 
    \item \textbf{{How quickly does performance degrade when the POMDP model used for planning does not match the real world?}}
        Since it is difficult to guarantee that models, especially human models, match the dynamics of the world exactly, it is important to understand the effects of model mismatch.
        This work characterizes performance when the parameters of the world simulation model do not match the planning model.
\end{enumerate}

The paper proceeds as follows: \Cref{sec:background} gives a short introduction to multi-objective optimization and POMDP planning. \Cref{sec:problem} gives a detailed problem description, and solution approaches are described in \Cref{sec:approaches}. Finally, \Cref{sec:results} addresses the questions above with numerical results.

\section{Background} \label{sec:background}

This section introduces some concepts and techniques used for planning in autonomous driving.

\subsection{Objectives for Automated Driving}

Autonomous vehicles are designed for multiple objectives.
Two of the most important are safety and efficiency.

\emph{Safety} is often defined as a guarantee that a certain set of negative outcomes will never occur, that is, unsafe states are \emph{unreachable} even under worst-case disturbances.
However, this notion of safety is clearly unattainable in many cases because other vehicles may easily force collisions if they take adversarial actions.
Thus, autonomous vehicle engineers must pursue a more modest objective: minimizing the likelihood of reaching unsafe states.

\emph{Efficiency} is accomplishing a goal with minimum resource use.
This can take a variety of forms in different contexts, for instance maximizing the throughput of cars on a highway or reducing energy consumption or emissions.
One of the most valuable resources that autonomous vehicles will manage is time; they must be judicious in their use of time to reach a destination.
In the particular scenario studied in this work, efficiency corresponds to the vehicle's ability to make several lane changes in a fixed distance.

Safety and efficiency have a complex relationship.
Usually they are opposed to one another.
For example, maintaining slower speeds and yielding to other drivers is typically safer but is also less efficient.
At first, it may seem that safety is a strictly higher priority than efficiency, but consumers will not sacrifice efficiency without limit \cite{moen2007determinants}; if an autonomous car gains a reputation of reaching its destination 5 minutes later than competitors, it is reasonable to assume that some vehicle owners or riders will choose another competitor.
Moreover, in some cases acting too cautiously in a way that is uncharacteristic of human drivers can also be unsafe \cite{naughton2015freeway,schoettle2015crashes}.

In addition to safety and efficiency, autonomous cars may need to take other factors like passenger comfort, profits for manufacturers and operators, environmental impact, and social utility into account.

\subsection{Pareto Frontier Analysis}

One approach to multiobjective problems encountered in autonomous driving is to create a single objective function that is a weighted sum of the functions corresponding to the individual objectives,
\begin{equation} \label{eq:scalarized}
    \underset{\pi}{\text{maximize}} \quad J_0(\pi) + \lambda_1 J_1(\pi) + \lambda_2 J_2(\pi) + \ldots + \lambda_n J_n(\pi) \text{,}
\end{equation}
where $\lambda_i$ are the relative weights.
For example in this work, $J_0$ will be an efficiency reward, and $J_1$ will be a safety reward.
This scalarization approach allows conventional decision-making algorithms to be used, but choosing appropriate weights before solving the problem is often difficult because there is not usually a straightforward mapping from meaningful performance metrics to weights~\cite{kochenderfer2019algorithms}.

This work adopts a stance that is agnostic to the value $\lambda$ by comparing Pareto frontiers. 
A solution is said to be \emph{Pareto-optimal} if no objective can be improved without adversely affecting another objective.
The \emph{Pareto frontier} is the set of all Pareto-optimal solutions.
Every solution to problem (\ref{eq:scalarized}) for a positive value of $\lambda$ lies on the convex hull of the Pareto front.
Additional Pareto-optimal points that are not part of the convex hull may also exist, but they cannot be found by solving problem (\ref{eq:scalarized})~\cite{kochenderfer2019algorithms}.

Since exact solutions to the problems studied in this work are intractable, it is impossible to reliably generate points on the true Pareto front.
Instead, approximate Pareto fronts are constructed by connecting approximate solutions to (\ref{eq:scalarized}) with straight lines (see \cref{fig:gaps} for examples).
Given two algorithms, A and B, if the Pareto front for A is in a better position than that of B for most or all values of $\lambda$, then it is possible to argue that algorithm A is superior to algorithm B without committing to a particular value of $\lambda$.

\subsection{POMDPs}

The partially observable Markov decision process (POMDP) is a mathematical formalism that can represent a wide range of sequential decision making problems~\cite{sutton2018reinforcement,kochenderfer2015decision}. 
In a Markov decision process (MDP), an agent takes \emph{actions} that affect the state of the system with the goal of maximizing a \emph{reward} collected by visiting certain states.
In a POMDP, this task is made much more difficult because the agent only gains information about the state through noisy \emph{observations}. 

A POMDP is defined by the tuple $(\sspace, \aspace, \tdist, \reward, \ospace, \odist, \gamma)$.
$\sspace$ is the state space, and
$\aspace$ is the action space. 
The transition model, $\tdist(s' \mid s, a)$, encodes the probability of transitioning to state $s'$ given that action $a$ is taken in state $s$.
The reward function, $\mathcal{R}(s, a, s')$, specifies the reward for transitioning from $s$ to $s'$ when action $a$ is taken.
$\ospace$ is the observation space, and $\odist$ is the observation model, where $\odist(o \mid s)$ is the probability or probability density of receiving observation $o$ in state $s$~\cite{kochenderfer2015decision,kaelbling1998planning}.
Finally, $\gamma \in [0,1]$ governs how reward is discounted in the future (see \cref{eq:value}).

Since an agent acting in a partially observable environment does not have perfect access to the state, it must make decisions based only on the history of observations that it has received.
Bayes rule can be used to calculate the probability that the system is in each state conditioned on the previous observations.
This distribution contains sufficient information to choose optimal actions~\cite{kochenderfer2015decision,kaelbling1998planning}, and is known as the \emph{belief}; $\mathcal{B}$ is the space of all possible beliefs.
Choosing actions based on the belief is often more straightforward than choosing based on the history, and the discussion in this work will be constrained to approaches that make decisions based on belief.

The objective in a POMDP is to find a policy, $\pi$, that maps a belief to an action maximizing the \emph{value function}, the expected sum of all future rewards, denoted with
\begin{equation} \label{eq:value}
    V^\pi(b) \equiv E\left[\left.\sum_{t=0}^{\infty} \gamma^t R(s_t, \pi(b_t), s_{t+1}) \right| b_0 = b \right]\text{,}
\end{equation}
where subscript $t$ denotes the time step.
The discussion in this work will also refer to the belief-action value function, $Q$.
The $Q$ value is the value of taking action $a$ at the current step and then executing a specified policy at future steps,
\begin{equation}
    Q^\pi(b, a) \equiv E\left[\left. R(s, a) + \gamma V^\pi(b') \right| b \right]
\end{equation}
The value function is important because it provides a basis for dynamic programming solutions, and a POMDP optimization problem can be written as
\begin{equation}
    \underset{\pi:\, \mathcal{B}\to\mathcal{A}}{\mathop{\text{maximize}}} \, V^\pi(b) \text{.}
\end{equation}
When the superscript $\pi$ is omitted, $V$ refers to the optimal value or an estimate of the optimal value.

In some cases, the solution to the underlying fully observable MDP corresponding to a POMDP is useful \cite{littman1995learning}. For a POMDP $(\sspace, \aspace, \tdist, \reward, \ospace, \odist, \gamma)$, the underlying MDP is $(\sspace, \aspace, \tdist, \reward, \gamma)$. The optimal state-action value function for this MDP will be denoted with
\begin{multline} \label{eq:qmdp}
    Q_\text{MDP}(s, a) = \\
    R(s, a) + \max_{\pi:\sspace \to \aspace} E\left[\left.\sum_{t=1}^\infty \gamma^t R(s_t, \pi(s_t)) \right| s_0 = s \right] \text{.}
\end{multline}

\subsection{Monte Carlo Tree Search} \label{sec:mcts}

Though POMDPs are very powerful in terms of expression, even the class of finite-horizon POMDPs is PSPACE-complete, indicating that it is unlikely that efficient general exact algorithms for large problems will be discovered \cite{papadimitriou1987complexity}.
Because of this, approximations are used.
In this work, five different approximate POMDP solution methods, described in detail in \cref{sec:planning}, are considered.
All of these approaches use variants of Monte Carlo tree search (MCTS)~\cite{browne2012mcts} with double progressive widening (DPW)~\cite{couetoux2011double}, but each makes a different set of assumptions.

Given an initial state (or belief in the POMDP case), a policy may be represented by a tree with alternating layers of state and action nodes in the MDP case or observation and action nodes in the POMDP case (see \cref{fig:approaches} for depictions of several approximate versions of such a tree).
MCTS uses Monte Carlo simulations to incrementally construct and search only important parts of this tree.

This technique reduces computation compared to an exact offline (PO)MDP solution (which is impossible for a problem as large as this) in several ways.
First, by computing actions \emph{online} starting from the current state, the solver must only consider states that are likely to be visited in the near future.
Second, by estimating the value function $Q(s, a)$ (or $Q(b, a)$ in the POMDP case) at each action node and running more simulations for actions that are promising according to an upper confidence bound~\cite{browne2012mcts}, it limits computation wasted on parts of the tree that are not likely to be in the optimal policy.

Double progressive widening further focuses computation by considering only a limited, but gradually increasing, number of sampled states or observations.
In this way, the solver is able to search deeper into the tree, that is, further into the future, than algorithms that consider more states.

In some of the approaches used in this work, MCTS is directly applicable because they involve solving an MDP that approximates the true POMDP.
However, in the full POMDP formulation, the state is not directly observed and planning must be conducted in the belief space.
Since full Bayesian belief updates are computationally expensive, the POMCPOW algorithm extends MCTS to include approximate beliefs represented by weighted particle collections that are gradually improved as the tree is searched~\cite{sunberg2018pomcpow}.
Though the particle weighting scheme has been shown to be sound~\cite{lim2020sparse}, a proof that POMCPOW converges to the optimal POMDP solution has not yet been given. The large number of trajectories that POMCPOW is able to consider using its weighted particle approach are shown in~\cref{fig:pomcpow}.

\subsection{Related Work}

Significant effort has been invested in identifying accurate human driver models for planning.
Specific thrusts include learning these models from data \cite{grindele2015trafficmodel,sadigh2014}, representing these models efficiently \cite{WheelerRobbelKochenderfer2015,WheelerKochenderfer2016}, recognizing the intent of other drivers online \cite{dc2015identifying}, using game-theoretic approaches to model interaction \cite{fridovich2019efficient,peters2020alignment}, and predicting motion with models that respect kinematics and inferred constraints \cite{luo2019gamma}.

Though sensors accurately measure much of the physical state of other vehicles, the internal state (e.g., intentions and aggressiveness) of other road users can only be indirectly inferred \cite{sadigh2016gathering, bai2015intention,lam2015improving,dc2015identifying,luo2019gamma,sunberg2017value}.
This fact is central to the hypothesis explored in this work: that inferring and planning with an estimate of the internal states of the traffic participants will improve safety and efficiency.

POMDPs are particularly well suited for modeling decisions for autonomous vehicles because they explicitly capture limitations of the vehicle's sensors in measuring relevant state variables. Some research focuses on \emph{physical} variables that are hidden because of occlusions or other sensor limitations \cite{brechtel2013mcvi,bouton2018scalable,lin2019decision,hubmann2019occlusions,ulbrich2013probabilistic}.
Other research focuses on latent states internal to human drivers \cite{sadigh2016gathering,bouton2017belief,hubmann2018automated,liu2015situation,meghjani2019context} or pedestrians \cite{bai2015intention,luo2018porca}.

In addition to MDP and POMDP planning, there are many other methods for planning lane changes~\cite{bevly2016lane}.
The experiments presented here are not meant to compare fundamentally different planning methods, but rather to contrast similar tree search methods using POMDP models that consider internal states and MDP models that do not.


\section{Problem Formulation} \label{sec:problem}

The hypothesis is tested in a freeway driving scenario (\cref{fig:pomcpow}).
A vehicle must navigate from the rightmost to the leftmost lane of a four lane freeway within a specified distance while maintaining safety and comfort.

The driver models used in this work are collision-free, that is, all human-driven and autonomous vehicles will stop or slow in time to avoid a collision.
The decision to adopt collision-free models was made because no suitable model for the interactions of human-driven vehicles leading up to a collision was found.
Thus instead of counting collisions to evaluate the safety of the system, any situation in which \emph{any} human-driven or autonomous vehicle has to break hard to avoid a collision is marked unsafe.
The following sections describe the mathematical details of the simulation model.

\subsection{POMDP Formulation}

The freeway driving scenario is formulated as a POMDP with the following elements:

\begin{itemize}
    \item State space, $\sspace$: A system state, $$s = \left(q_0, \phys_1,\beh_1, \ldots \phys_N, \beh_N \right) \in \sspace \subset \reals^{124}\text{,}$$ consists of the physical state of the ego vehicle ($q_0$), and physical state and behavior model for each of the $N$ other cars in the scene.
    A physical state, $\ith{\phys} = (\ith{x},\ith{y},\ith{\dot{x}},\ith{\dot{y}}) \in \reals^4\text{,}$ consists of the car's longitudinal and lateral position and velocity. An internal state, $\ith{\beh} \in \reals^8$, consists of values for the behavior parameters listed in \cref{tab:modelparams}. Since the number of vehicles on the road is limited to $N_\text{max}=10$, the full state including the ego physical state can have up to 124 dimensions.
    \item Action space, $\aspace$: An action, $a = (\ego{\ddot{x}}, \ego{\ddot{y}}) \in \aspace$, consists of the longitudinal acceleration and lateral velocity of the \av{} vehicle. The action space is discrete and pruned to prevent crashes (see \cref{sec:action}).
    \item State transition model, $\tdist$: The state transition distribution is implicitly defined by a generative model that consists of a state transition function, $F(\cdot)$, and a stochastic noise process (see \cref{sec:dynamics}).
    \item Reward model, $\reward$: The reward function, defined in \cref{sec:reward}, rewards reaching the left lane within the distance limit and penalizes unsafe actions.
    \item Observation space, $\ospace$: An observation, $o \in \ospace$ consists of the physical states of all of the vehicles, that is $o=(q_1, \ldots, q_N)$. No information about the internal state is directly included in the observation.
    \item Observation model, $\odist$: In these experiments the observation is a perfect measurement of the physical state, though the solution methods below can be used with noisy observations.
\end{itemize}
The remainder of this section elaborates on this model.

\subsection{Driver Modeling} \label{sec:driver}

The driver models for each car have two components: an acceleration model that governs the longitudinal motion and a lane change model that determines the lateral motion.
In this work, the acceleration model is the Intelligent Driver Model (IDM) \cite{treiber2000idm}, and the lane change model is the ``Minimizing Overall Braking Induced by Lane change'' (MOBIL) model \cite{kesting2007mobil}.
Both have a small number of parameters that determine the  behavior of the drivers.
The distribution of these parameters in the population of vehicles will be denoted $\Beh$.

The IDM Model was developed as a simple model for ``microscopic'' simulations of traffic flows and is able to reproduce some phenomena observed in real-world traffic flows~\cite{treiber2000idm}.
It determines the longitudinal acceleration for a human-driven car, $\ddot{x}$, based on the desired distance gap to the preceding car, $g$, the absolute velocity, $\dot{x}$, and the velocity relative to the preceding car $\Delta \dot{x}$.
The longitudinal acceleration is governed by the following equation:
\begin{equation}
    \ddot{x}_\text{IDM} = \accelparam \left[ 1 - \left( \frac{\dot{x}}{\dot{x}_0} \right)^{\delta} - \left(\frac{g^*(\dot{x}, \Delta \dot{x})}{g}\right)^2 \right] \text{,}
\end{equation}
where $g^*$ is the desired gap given by
\begin{equation} \label{eqn:gstar}
    g^*(\dot{x}, \Delta \dot{x}) = g_0 + T \dot{x} + \frac{\dot{x}\Delta \dot{x}}{2 \sqrt{\accelparam b}} \text{.}
\end{equation}
Brief descriptions and values for the parameters not defined here are provided later in \cref{tab:modelparams}.

A small amount of noise is also added to the acceleration
\begin{equation}
    \ddot{x} = \ddot{x}_\text{IDM} + w \text{,}
\end{equation}
where $w$ is a random variable with a triangular distribution with support between $-\accelparam/2$ and $\accelparam/2$. In cases where the noise might cause a hard brake or lead to a state where a crash is unavoidable, the distribution is scaled appropriately.

The MOBIL model makes the decision to change lanes based on maximizing the acceleration for the vehicle and its neighbors.
When considering a lane change, MOBIL first ensures that the safety criterion $\tilde{\ddot{x}}_n \geq -b_\text{safe}$, where $\tilde{\ddot{x}}_n$ will be the acceleration of the following car if the lane change is made and $b_\text{safe}$ is the safe braking limit. 
It then makes the lane change if the following condition is met
\begin{equation}
    \tilde{\ddot{x}}_c - \ddot{x}_c + p \left( \tilde{\ddot{x}}_n - \ddot{x}_n + \tilde{\ddot{x}}_o - \ddot{x}_o \right) > \Delta a_\text{th}
\end{equation}
where the quantities with tildes are calculated assuming that a lane change is made, the quantities with subscript $c$ are quantities for the car making the lane change decision, those with $n$ are for the new follower, and those with $o$ are for the old follower.
The parameter $p \in [0,1]$ is the politeness factor, which represents how much the driver values allowing other vehicles to increase their acceleration. The parameter $\Delta a_\text{th}$ is the threshold acceleration increase to initiate a lane changing maneuver. Parameter values are listed in \cref{tab:modelparams}.

\subsection{Physical Dynamics} \label{sec:dynamics}

The physical dynamics assume time is divided into discrete steps of length $\dt$.
The dynamics assume constant longitudinal acceleration and constant lateral velocity over a time step. 

There is a physical limit to the braking acceleration, $\bmax$.
Lateral velocity is allowed to change instantly because cars on a freeway can achieve the lateral lane change velocities in time much shorter than $\dt$ by steering.
If MOBIL determines that a lane change should be made, the lateral velocity, $\dot{y}$, is set to $\dot{y}_\text{lc}$.
Lane changes may not reverse; once a lane change has begun, $\dot{y}$ remains constant until it is complete (this is the reason that $\dot{y}$ is part of the state).
When a vehicle reaches the midpoint of a lane, lateral movement stops so that lane changes always end at lane centers.

Since MOBIL only considers adjacent lanes, there must be a coordination mechanism so that two cars do not converge into the same lane simultaneously.
If two cars begin changing into the same lane simultaneously and the front vehicle is within $g^*$ of the rear vehicle, the rear vehicle's lane change is canceled.

Because vehicles far from the ego are less relevant, the scene is limited to \SI{50}{\meter} in front of and behind the \av{} vehicle. 
Thus, a model for vehicle entry into this section is needed.
If there are fewer than $N_\text{max}$ vehicles on the road, a new vehicle is generated.
First, a behavior for the new vehicle is drawn from $\Beh$ with initial speed $\dot{x}_0 + \sigma_\text{vel} w_0$, where $\dot{x}_0$ is the desired speed from the behavior model and $w_0$ is a zero-mean, unit-variance, Gaussian random variable independent for each car.
If this speed is greater than the \av{}'s speed, the new vehicle will appear at the back of the road section; if it is less, it will appear at the front.
For each lane, $g^*$ is calculated, either for the new vehicle if the appearance is at the back or for the nearest following vehicle if the appearance is at the front.
The new vehicle appears in the lane with the largest clearance.
If no clearance is greater than $g^*$, the new vehicle does not appear.
Once the \av{} reaches the target lane ($y = y_\text{target}$) or passes the distance limit ($x \geq L$), the problem terminates.

Throughout this paper, the behavior described so far will be denoted compactly by the state transition function
\begin{equation}
    s' = F(s,u,w) \, \text{.}
\end{equation}

\subsection{Action Space for Crash-Free Driving} \label{sec:action}

At each time step, the planner for the \av{} must choose the longitudinal acceleration and lateral speed.
For simplicity, the vehicle chooses from up to ten discrete actions.
The vehicle may make an incremental decrease or increase in speed or maintain speed, and it may begin a left or right lane change or maintain the current lane.
The combination of these adjustments make up nine of the actions.
The final action is a braking action determined dynamically based on the speed and position of the vehicle ahead.
In most cases, the acceleration for this action is a nominal value, $-b_\text{nominal}$, but this is sometimes overridden.
At each time step, the maximum permitted acceleration, $a_\text{max}$, is the maximum acceleration that the \av{} could take such that, if the  vehicle ahead immediately begins braking at the physical limit, $\bmax$, to a stop, the \av{} will still be able to stop before hitting it without exceeding physical braking limits itself.
The braking action is $(\ego{\ddot{x}}, \ego{\ddot{y}}) = (\min \{a_\text{max}, -b_\text{nominal} \}, 0)$.


The dynamic braking action guarantees that there will always be an action available to the \av{} to avoid a crash.
At each step, the action space is pruned so that if $\ego{\ddot{x}} > a_\text{max}$ or if a lane change leads to a crash, that action is not considered.
Since the IDM and MOBIL models are both crash-free \cite{kesting2009agents}, and actions that lead to crashes for the \av{} are not considered, no crashes occur in the simulation.
Eliminating crashes is justifiable because it is likely that an actual autonomous vehicle would have a low-level crash prevention system to increase safety and facilitate certification. 
In addition, modelling driver behavior in the extraordinary case of a crash is difficult. 

\subsection{Reward Function and Objectives} \label{sec:reward}

The broad goals of efficiency and safety manifest as two concrete objectives in the lane change problem: reaching the target lane within a specified distance, $L$, (efficiency) and maintain the comfort and safety of both the \av{} and the other nearby vehicles (safety).
Thus, the following two metrics are used to evaluate planning performance: 1) the fraction of episodes in which the \av{} reaches the target lane, and 2) the fraction of episodes in which \emph{any} vehicle operates unsafely.

For this work, hard braking and unusually slow velocity are considered unsafe\footnote{There are many safety surrogate measures that could be used in this context.
We expect the broad conclusions of this study should hold for a variety of safety measures, and these metrics are often closely related.
For example, there is a special relationship to the commonly used time-to-collision metric.
In particular, if the time-to-collision under assuming $\ddot{x} = -b_\text{hard}$ is less than one time step, the vehicle will be forced to brake hard.}.
A hard braking maneuver is defined as $\ddot{x} < -b_\text{hard}$ and slow velocity as $\dot{x} < \dot{x}_\text{slow}$, where $b_\text{hard}$ and $\dot{x}_\text{slow}$ are chosen to be uncomfortably abrupt deceleration or slow travel that might result in an accident in real conditions (see \cref{tab:params}).
In addition to quantifying safety, hard braking also serves as a proxy for comfort.

In order to encourage the planner to maximize these metrics, the reward function for the POMDP is defined as follows: 
\begin{equation} \label{eqn:reward}
\begin{split}
    \reward(s, a, s') =& \text{ in\_goal}(s')\\
                       & - \lambda \text{ any\_hard\_brakes}(s, s')\\
                       & - \lambda \text{ any\_too\_slow}(s')
\end{split}
\end{equation}
where
\begin{align*}
    \text{in\_goal}(s') &= \mathbf{1}(\ego{y}' = y_\text{target}, \ego{x}' \leq L) \text{,}\\
    \text{any\_hard\_brakes}(s,s') &= \max_{i \in \{1, \ldots, N\}}\{\mathbf{1}(\ith{\dot{x}}' - \ith{\dot{x}} < -b_\text{hard} \Delta t) \} \text{,}\\
    \text{any\_too\_slow}(s') &= \max_{i \in \{1, \ldots, N\}} \{\mathbf{1}(\ith{\dot{x}'} < \dot{x}_\text{slow})\} \text{.}
\end{align*}
That is, there is a positive reward for reaching the target lane within the distance limit, and hard brakes and slow velocity for any car are penalized.
The weight $\lambda$ balances the competing goals and can be used to create an approximate Pareto frontier.
 


\section{Solution Approaches} \label{sec:approaches}

The planning approaches investigated in this work all involve approximate planning with variations of MCTS-DPW (\cref{sec:mcts}).
These variations are enumerated and described below. 
For some of these variations, a belief over the internal states of other drivers is maintained with a particle filter which is described in detail in \cref{sec:filtering}.

\subsection{Approximate Planning Approaches} \label{sec:planning}

\begin{figure*}[tbp]
    \vspace{1mm} 
    \centering
    \subfloat[][Assume normal behavior]{
        \includegraphics{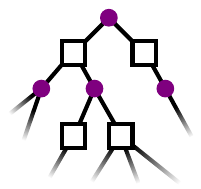}
    }
    \subfloat[][Naive MDP]{
        \includegraphics{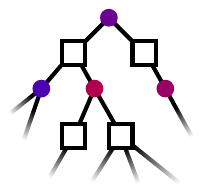}
    }
    \subfloat[][Mean state MDP]{
        \includegraphics{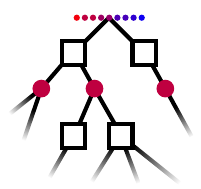}
    }
    \subfloat[][QMDP]{
        \includegraphics{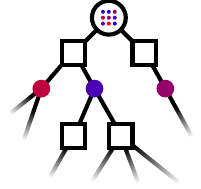}
    }
    \subfloat[][POMCPOW]{
        \includegraphics{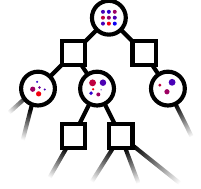}
    }
    \caption{Comparison of approximate planning approaches. Squares correspond to action nodes, solid circles to states or state nodes, with different colors representing different internal states of the other drivers, and unfilled circles to belief nodes.}
    \label{fig:approaches}
    \vspace{-1em}
\end{figure*}

Five approximate planning variations were used. They are enumerated below and illustrated in \cref{fig:approaches}.
\begin{enumerate}[wide]
    \item \textbf{Assume normal behavior}: The first approach uses MCTS-DPW to solve an MDP assuming that all drivers act with identical ``normal'' internal state (see \cref{tab:modelparams}).
    \item \textbf{Model all uncertainty as aleatoric uncertainty (Naive MDP)}: This approach uses MCTS-DPW to plan as if all uncertainty is simply aleatoric uncertainty, that is, as if the problem were an MDP with a state consisting only of the physical state and the internal states were random variables, independent at each timestep, distributed according to the internal state distribution $\Beh$. 
This model would be the result of fitting a Markov model with only the physical state based on data from all drivers.

    \item \textbf{Mean state MDP (MSM)}: In this approach, a particle filter is used to maintain a belief distribution over the driver internal states. At each timestep MCTS-DPW is used to solve the fully observable MDP using the mean internal state values $E_{s \sim b}[s]$. This approach is sometimes called certainty equivalence control~\cite{mania2019certainty}. Here MCTS approximates the policy
    \begin{equation}
        \pi_\text{MSM}(b) = \underset{a}{\text{argmax}}\,Q_\text{MDP}\left(\underset{s \sim b}{E}[s], a\right) \text{.}
    \end{equation}

    \item \textbf{QMDP}: 
        The QMDP approximation is based on an algorithm proposed by \citet{littman1995learning} that uses value iteration to find the $Q_\text{MDP}$ values (\cref{eq:qmdp}) for a POMDP and then takes the action that maximizes the expected $Q_\text{MDP}$ value for the belief.
        In this work, the $Q_\text{MDP}$ values are estimated through MCTS-DPW instead of value iteration.
        The expected $Q_\text{MDP}$ value for a belief action pair is 
        \begin{align}
            Q_\text{MDP}(b,a) &= \underset{s \sim b}{E}\left[Q_\text{MDP}(s,a)\right] \\
                              &= \int_{s \in \sspace} Q_\text{MDP}(s,a) b(s) \, ds \text{,}
        \end{align}
        and MCTS-DPW seeks to find the policy
        \begin{equation}
            \pi_\text{QMDP}(b) = \underset{a}{\text{argmax}} \underset{s \sim b}{E}\left[Q_\text{MDP}(s,a)\right]\text{.}
        \end{equation}
        Since it chooses the action that maximizes this value, it is easy to see that the QMDP approximation is the optimal solution to a hypothetical problem with partial observability on the current step, but that subsequently becomes fully observable.

    \item \textbf{POMCPOW}: This approach uses the POMCPOW algorithm~\cite{sunberg2018pomcpow} that leverages weighted particle filtering as described in \cref{sec:mcts} to find an approximate solution to the POMDP. \Cref{fig:pomcpow} showcases this technique. 
\end{enumerate}

The first two approaches are baselines, representing ways to force the epistemic uncertainty in the POMDP into an aleatoric-only MDP formulation.
The first optimistically assumes that it knows the internal states of other drivers, making it an overconfident baseline.
The second, on the other hand, is a conservative baseline; it plans pessimistically assuming it can learn nothing new about the drivers.

The MLMDP and QMDP methods passively learn online using particle filtering.
However, during planning, they assume that all information about the state of the problem is known (MLMDP) or will become known after one step (QMDP).
This makes these planners overly optimistic and thus systematically suboptimal.
Moreover, because of this optimistic assumption about knowledge during planning, there is no incentive for learning about the state, and hence these policies will not take costly actions for active learning, even if such actions are part of the optimal solution to the POMDP.
Nevertheless, these approximations are useful in many domains.

POMCPOW is the closest approximation to the exact POMDP solution~\cite{sunberg2018pomcpow,lim2020sparse}. It considers epistemic state uncertainty at deeper levels of the tree, and is thus able to find the best policies for this problem.

All of these MCTS planning methods are limited to 1000 iterations, which takes approximately \SI{0.5}{\second} on a single core CPU, allowing the algorithms to run faster than real time with $\Delta t = \SI{0.75}{\second}$. The computational load scales linearly with the number of vehicles, so the algorithm could be used for moderately larger scenarios and slightly modified versions could be further scaled with parallelization. The value at new leaf nodes is estimated with a simple rollout policy that moves toward the goal lane if a lane change is safe and otherwise accelerates or decelerates depending on whether there is more distance to the front or rear.

\subsection{Internal State Filtering} \label{sec:filtering}

In the MLMDP, QMDP, and POMCPOW approaches, online estimation of $\beh$ is accomplished with a particle filter~\cite{thrun2005probabilistic}.
Filtering is independent for each car, but all of the behavior parameters for a given car are estimated jointly.
There are two versions of the filter.
In the first, a particle, $\hat{\theta}$, consists of values of all model parameters.
In the second, all parameters are assumed perfectly correlated (see \cref{sec:dist}), so a particle consists of only a single value, the ``aggressiveness''.

The belief at a given time consists of the exactly known physical state, \phys, and a collection of $M$ particles, $\{\hat{\theta}^k\}_{k=1}^M$, along with associated weights, $\{W^k\}_{k=1}^M$.
To update the belief when action $u$ is taken, $M$ new particles are sampled with probability proportional to the weights, and sampled noise values $\{\hat{w}^k\}_{k=1}^M$ are used to generate new states according to ${\hat{s}^k}{'} = F((\phys, \hat{\theta}^k), u, \hat{w}^k)$.
The new weights are determined by approximating the conditional probability of the particle given the observation:
\begin{align*}
    {W^k}{'} &= \left.
\begin{cases}
    \max\left\{0, \frac{a - 2 \, \left| \dot{x}' - \hat{\dot{x}}' \right|}{a} \right\} & \text{if } y' = \hat{y}' \\
\gamma_\text{lane} \max\left\{0, \frac{a - 2 \, \left| \dot{x}' - \hat{\dot{x}}' \right|}{a} \right\} & \text{o.w.}
\end{cases} \right\} \\
    & \appropto \operatorname{Pr}\left(\left.\hat{\theta}^k \right| o \right)
\end{align*}
where $\dot{x}'$ and $y'$ are from the observation, $\hat{\dot{x}}'$ and $\hat{\dot{y}}'$ are from ${\hat{s}^k}{'}$, the max expression is proportional to the density of the acceleration noise triangular distribution, and $\gamma_\text{lane} \in [0,1]$ is a hand-tuned penalty for incorrect lane changes (see \cref{tab:params}).

\section{Results} \label{sec:results}

The computational results from this study are designed to meet the two goals of 1) quantifying the size of the gap between the baseline control algorithm and the maximum potential lane change performance and 2) showing which cases internal state estimation and POMDP planning can approach the upper bound on performance.

Experiments are carried out in three scenarios, each with a different distribution of internal states.
In each of these scenarios, each of the approaches described in \cref{sec:approaches} are compared with an approximate upper performance bound obtained by planning with perfect knowledge of the behavior models.
Initial scenes for the simulations are generated by beginning a simulation with only the \av{} on the road section and then simulating 200 steps to allow other vehicles to accumulate.
The open source code for these experiments can be found at \url{https://github.com/sisl/Multilane.jl/tree/master/thesis}.

\subsection{Driver Model Distribution Scenarios} \label{sec:dist}

For the numerical testing, three internal state distribution scenarios are considered.
In all of these scenarios, drivers behave according to the models presented in \cref{sec:driver}, however the IDM and MOBIL parameter values are distributed differently in each scenario.

The values are drawn from a continuous joint distribution, but \cref{tab:modelparams} shows typical parameter values for aggressive, timid, and normal drivers.
The values are taken from \citet{kesting2009agents}, but some have been adjusted slightly so that the normal driver parameters are exactly half way between the timid and aggressive values.
In all three of the scenarios, the \emph{marginal} distributions of the parameters are uniformly distributed between the aggressive and timid values.

The difference between the scenarios is in the correlation of the parameter values.
In this context, ``correlation'' does not refer to correlation between different drivers in the population, i.e. correlation between $\beh_i$ and $\beh_j$ where $j \neq i$.
Rather, it refers to correlation between different parameters within individual driver models, i.e. correlation between the parameters such as $T$ and $\dot{x}_0$ within a driver model $\beh_i$.
Each driver model, $\beh_i$, is sampled \emph{independently} from a joint parameter distribution, $\Beh$, that is identical across all drivers, but we study various levels of correlation between the individual parameters.
Intuitively, a distribution with high correlation corresponds to the assumption that nearly all polite drivers (high $p$) also prefer slower speed (low $\dot{x}_0$) and timid values for all parameters in \cref{tab:modelparams}.
Conversely, low correlation corresponds to the assumption that polite lane changers prefer a range of speeds and other parameter values.

In order to maintain consistent marginal distributions while varying the correlation, copulas are used.
A copula is a multivariate distribution with uniform marginal distributions but a joint distribution with correlation between the variables \cite{nelsen2007introduction}.
An $n$-dimensional Gaussian copula is defined by a correlation matrix, $\Sigma$, and has the cumulative distribution function (CDF)
\begin{equation}
    F_{\text{GC}}(x) = \Phi_\Sigma (\Phi^{-1}(x_1), \ldots, \Phi^{-1}(x_n))\text{,}
\end{equation}
where $\Phi_\Sigma$ is the CDF for a multivariate Gaussian distribution with covariance $\Sigma$, and $\Phi^{-1}$ is the inverse CDF for a univariate Gaussian distribution.
If random vector $X$ has CDF $F_{\text{GC}}$, then for any $i \neq j$, the correlation between $X_i$ and $X_j$ is $\Sigma_{ij}$.

In Scenario 1, all of the parameters are independently distributed.
In Scenario 2, all of the parameters are perfectly correlated so that all parameters are deterministic functions of a single uniformly-distributed random variable, the ``aggressiveness'' of the driver.
In Scenario 3, the distribution is correlated between these two extremes.
Specifically, it is a Gaussian copula with a covariance matrix
with 1 along the diagonal and correlation parameter $\rho$ elsewhere.
The values drawn from this distribution are scaled and translated to lie between the aggressive and normal limits.

For Scenario 3, the value of $\rho$ is \num{0.75}, and Scenarios 1 and 2 can be thought of as limiting cases where $\rho$ approaches 0 and 1, respectively.
In Scenarios 1 and 3, the first version of the particle filter, which estimates all of the model parameters jointly, is used, whereas in Scenario 2, the second version of the particle filter that assumes fully correlated parameters is used, that is, it only estimates a single ``aggressiveness'' parameter for each car.
The mean state MDP planner uses this ``aggressiveness'' parameter for all scenarios because this resulted in better performance.
The small scatter plots in \cref{fig:gaps} illustrate the level of correlation by plotting sampled values of two of the parameters.

\begin{table}[tbph]
    \caption{IDM and MOBIL parameters and extreme values.}
    \centering
    \footnotesize
    \begin{tabular}{@{}llrrr@{}}
        \toprule
        IDM Parameter & & \hspace{-3ex} Timid & Normal & Aggressive \\
        \midrule
        Desired speed (\si{\meter\per\second}) & $\dot{x}_0$ & 27.8 & 33.3 & 38.9 \\
        Desired time gap (\si{\second}) & $T$ & 2.0  & 1.5 & 1.0 \\
        Jam distance (\si{\meter})     & $g_0$ & 4.0  & 2.0 & 0.0 \\
        Max acceleration (\si{\meter\per\second\squared}) & $\bar{a}$ & 0.8  & 1.4 & 2.0 \\
        Desired deceleration (\si{\meter\per\second\squared}) & $b$ & 1.0  & 2.0 & 3.0 \\
        \midrule
        MOBIL Parameter & & \hspace{-3ex} Timid & Normal & Aggressive \\
        \midrule
        Politeness & $p$ & 1.0 & 0.5 & 0.0 \\
        Safe braking (\si{\meter\per\second\squared}) & $b_\text{safe}$ & 1.0 & 2.0 & 3.0 \\
        Acceleration threshold (\si{\meter\per\second\squared}) & $a_\text{thr}$ & 0.2 & 0.1 & 0.0 \\
        \bottomrule
    \end{tabular}
    \label{tab:modelparams}
\end{table}

\begin{table}[tbph]
    \caption{Various simulation parameters}
    \center
    \begin{tabular}{@{}llr@{}}
        \toprule
            Parameter & Symbol & Value \\
        \midrule
            Simulation time step & $\Delta t$ & \SI{0.75}{\second} \\
            Max other vehicles on road & $N_\text{max}$ & \num{10} \\
            Lane change rate & $\dot{y}_\text{lc}$ & \SI{0.67}{lanes\per\second} \\
            Distance limit & $L$ & \SI{1000}{\meter} \\
            Velocity noise standard deviation & $\sigma_\text{vel}$ & \SI{0.5}{\meter\per\second} \\
            Physical braking limit & $b_\text{max}$ & \SI{8.0}{\meter\per\second\squared} \\
            Penalized hard braking limit & $b_\text{hard}$ & \SI{4.0}{\meter\per\second\squared} \\
            Penalized minimum speed & $\dot{x}_\text{slow}$ & \SI{15}{\meter\per\second} \\
            UCT exploration parameter & $c$ & 8 \\
            DPW linear parameter & $k$ & 4.5 \\
            DPW exponent parameter & $\alpha$ & 0.1 \\
            MCTS search depth & & 40 \\
            MCTS iterations per step & & 1000 \\
            Particle filter wrong lane factor & $\gamma_\text{lane}$ & 0.05 \\
            Number of Particles (Joint Parameter Filter) & $M$ & 5000 \\
            Number of Particles (Aggressiveness Filter) & $M$ & 2000 \\
            Reward ratios for Pareto points & $\lambda$ & 0.5, 1, 2, 4, 8 \\
        \bottomrule
    \end{tabular}
    \label{tab:params}
\end{table}

\subsection{Pareto Front Comparison}

\begin{figure}
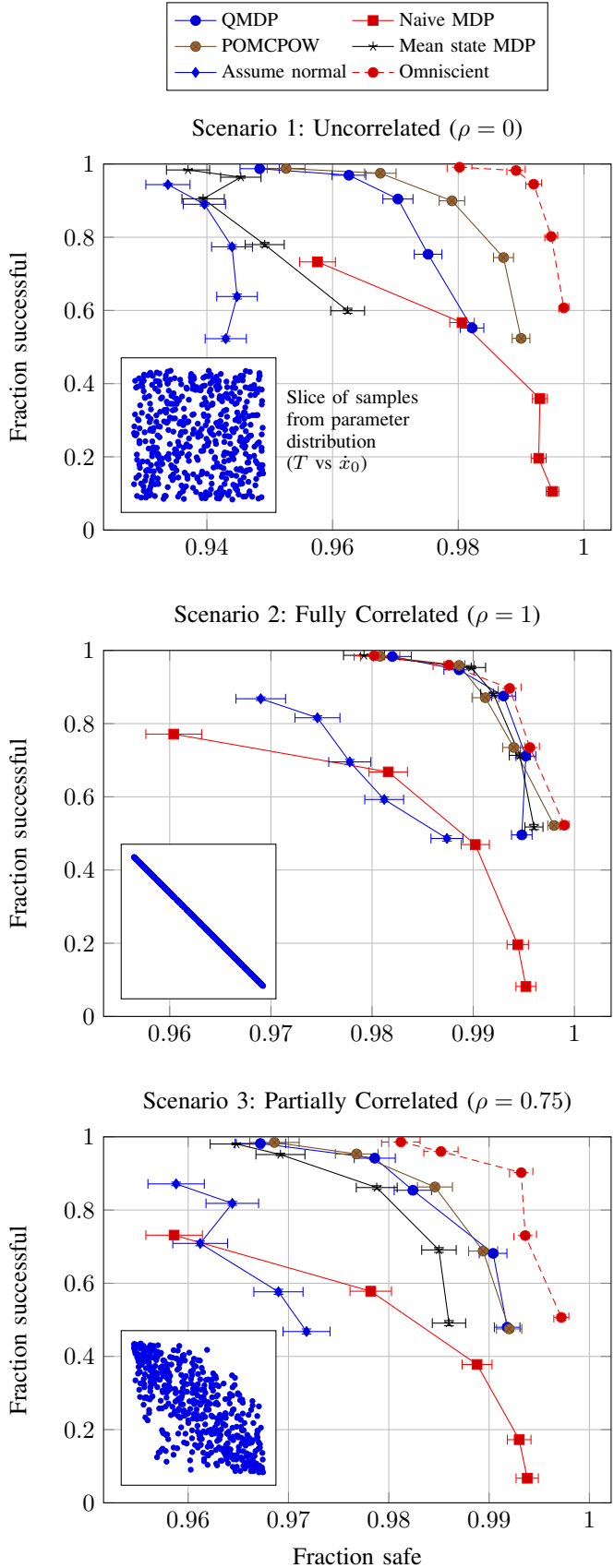

    \centering
    \input{gaps_000.tex}\\
    \vspace{5mm}
    \input{gaps_100.tex}\\
    \vspace{5mm}
    \input{gaps_075.tex}
    \vspace{-5mm}
    \caption{Approximate Pareto performance curves. Error bars indicate the standard error of the mean. The inset scatter plots indicate the correlation between parameters. Each point in the inset plot indicates a value of $T$ ($x$ axis) and $\dot{x}_0$ ($y$ axis) sampled from the parameter distribution.} \label{fig:gaps}
\end{figure}

\Cref{fig:gaps} shows approximate Pareto fronts illustrating the performance in terms of safety and efficiency of each of the approaches described in \cref{sec:approaches}.
Each of the points on the curve shows the result of \num{5000} independent simulations of the scenario with a particular safety-efficiency tradeoff weight, $\lambda$.

The baseline and upper bound approaches perform as expected.
The baseline planner that assumes all vehicles act with normal behavior parameters creates over-confident plans.
That is, it is able to reach the goal a large proportion of the time, but it causes many safety violations.
On the other hand, the naive MDP approximation is over-cautious. 
That is, it can attain a high level of safety, but it is never able to meet the goal more than \SI{80}{\percent} of the time.
The omniscient upper bound planner achieves performance equal to or greater than all other approaches.

As expected, better plans are attained as uncertainty is modeled more accurately.
The mean state MDP approach usually performs better than the baselines because it dynamically estimates the model parameters, but it is still overconfident (achieving a high success rate, but sacrificing safety) because it plans without any internal state uncertainty. 
QMDP performs better than mean state MDP because it considers samples from the entire estimated internal state distribution when planning.
Since the vehicle does not have to take costly information-gathering actions to accomplish its goal, POMCPOW only outperforms QMDP in certain cases.


One immediate concern that should be raised about the approximate Pareto frontiers in \cref{fig:gaps} is that they are not all convex.
The Pareto-optimal points generated by solving optimization problems of the form in (\ref{eq:scalarized}) must lie on the convex hull of the true Pareto front.
Thus, approximate Pareto fronts plotted by connecting particular solutions with straight lines as in \cref{fig:gaps} should be convex.
Particularly egregious violations of convexity can be found in the mean state MDP curve in \cref{fig:gaps}, Scenario 1, and the normal behavior assumption curve in \cref{fig:gaps}, Scenario 3, where there are ``kinks'' at the third point from the top ($\lambda=2$) that prevent these curves from even being monotonic.
The lack of convexity may be due to some combination of the following reasons:%
\begin{enumerate}[wide]
    \item The performance objectives plotted in the graphs do not exactly match the stage-wise reward function (\ref{eqn:reward}). For example, the planner observes a larger penalty if there are multiple safety violations, but this is not reflected in the plots.
    \item The MCTS-DPW solution method is itself stochastic and has no guarantees of convergence in finite time.
    \item Even given infinite computing time, the solvers will converge to inaccurate approximations of the true POMDP solution (except, perhaps, for POMCPOW).
\end{enumerate}

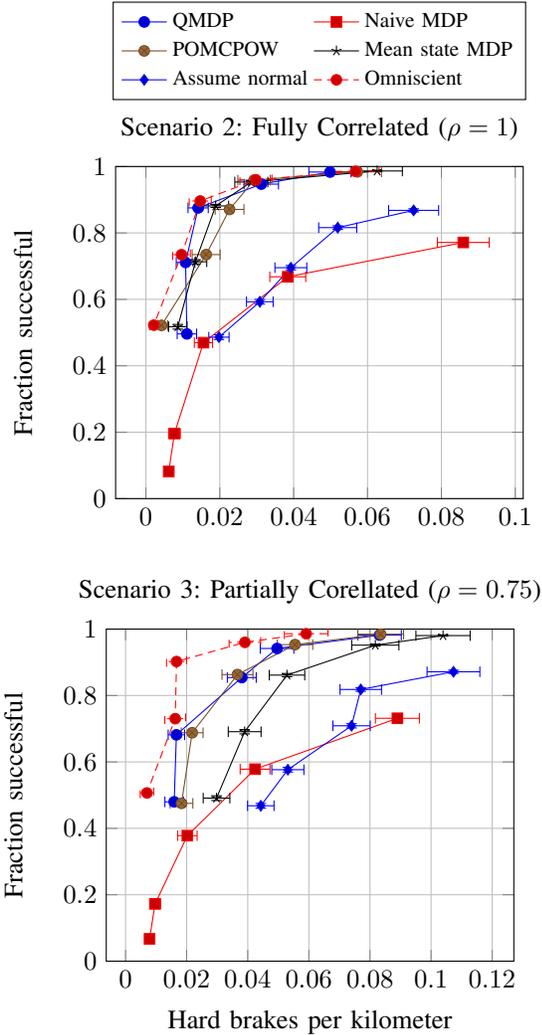
\begin{figure}
    \begin{center}
        \begin{tikzpicture}[]
\begin{axis}[height = {6cm},
    title = {Scenario 2: Fully Correlated ($\rho=1$)},
    ylabel = {Fraction successful}, ymax = {1.0},
    legend style={at={(0.5,1.2)}, anchor=south, font=\footnotesize}, legend columns=2, grid=both, xticklabel style={/pgf/number format/fixed, /pgf/number format/precision=2}, ymin = {0.0}, width = {7cm}]\addplot+ [
, error bars/.cd, 
x dir=both, x explicit, y dir=both, y explicit]
table [
x error plus=ex+, x error minus=ex-, y error plus=ey+, y error minus=ey-
] {
x y ex+ ex- ey+ ey-
0.04982981000707848 0.9834 0.005706248391264743 0.005706248391264743 0.0018070808250384592 0.0018070808250384592
0.031232783596988736 0.947 0.004668205805406493 0.004668205805406493 0.0031686287320555903 0.0031686287320555903
0.014140495513949719 0.8754 0.0027387047291886337 0.0027387047291886337 0.004671116768651089 0.004671116768651089
0.010740007530656943 0.7112 0.0025248322247608855 0.0025248322247608855 0.006409924291743685 0.006409924291743685
0.01109744050066768 0.496 0.0026314249494763134 0.0026314249494763134 0.0070715487243087185 0.0070715487243087185
};
\addlegendentry{QMDP}
\addplot+ [
, error bars/.cd, 
x dir=both, x explicit, y dir=both, y explicit]
table [
x error plus=ex+, x error minus=ex-, y error plus=ey+, y error minus=ey-
] {
x y ex+ ex- ey+ ey-
0.08592815286543948 0.7714 0.0070257612602544935 0.0070257612602544935 0.005939315035643325 0.005939315035643325
0.03840643277900424 0.668 0.00486665445597094 0.00486665445597094 0.006660636066815469 0.006660636066815469
0.015581010628548301 0.4696 0.00246857020717764 0.00246857020717764 0.007058692019532005 0.007058692019532005
0.007725156902563876 0.196 0.0015844950508215265 0.0015844950508215265 0.005614544026091883 0.005614544026091883
0.006186411204866511 0.0818 0.0013441969855230165 0.0013441969855230165 0.003876178137210714 0.003876178137210714
};
\addlegendentry{Naive MDP}
\addplot+ [
, error bars/.cd, 
x dir=both, x explicit, y dir=both, y explicit]
table [
x error plus=ex+, x error minus=ex-, y error plus=ey+, y error minus=ey-
] {
x y ex+ ex- ey+ ey-
0.057250781704335126 0.984 0.006528764780903003 0.006528764780903003 0.0017746633162313392 0.0017746633162313392
0.02927402353166062 0.959 0.004381808918702063 0.004381808918702063 0.002804527246188737 0.002804527246188737
0.02264634746138381 0.8708 0.0038952692679507356 0.0038952692679507356 0.0047440460784692095 0.0047440460784692095
0.016312281093638848 0.7348 0.0037838063718228605 0.0037838063718228605 0.006243523710026526 0.006243523710026526
0.004209628232537873 0.5216 0.0016879515277842213 0.0016879515277842213 0.007065173128400112 0.007065173128400112
};
\addlegendentry{POMCPOW}
\addplot+ [
, error bars/.cd, 
x dir=both, x explicit, y dir=both, y explicit]
table [
x error plus=ex+, x error minus=ex-, y error plus=ey+, y error minus=ey-
] {
x y ex+ ex- ey+ ey-
0.06260483925695728 0.9866 0.006835162140577933 0.006835162140577933 0.0016262278202591003 0.0016262278202591003
0.02854114963264953 0.9534 0.004472363049810522 0.004472363049810522 0.0029811852497117165 0.0029811852497117165
0.018891089897816835 0.8814 0.003499161204501022 0.003499161204501022 0.004572853616511231 0.004572853616511231
0.013411593483221355 0.7132 0.003059405212297941 0.003059405212297941 0.006396665964945574 0.006396665964945574
0.008640121551458229 0.5176 0.002494154850621593 0.002494154850621593 0.0070673925607342486 0.0070673925607342486
};
\addlegendentry{Mean state MDP}
\addplot+ [
, error bars/.cd, 
x dir=both, x explicit, y dir=both, y explicit]
table [
x error plus=ex+, x error minus=ex-, y error plus=ey+, y error minus=ey-
] {
x y ex+ ex- ey+ ey-
0.07248586845463478 0.868 0.00677120845795259 0.00677120845795259 0.004787461118044853 0.004787461118044853
0.05193311339416217 0.816 0.005151792492112015 0.005151792492112015 0.0054804020802667695 0.0054804020802667695
0.03924644806450444 0.6954 0.004313148197834641 0.004313148197834641 0.006509396473444352 0.006509396473444352
0.030837184686467223 0.5928 0.0036046152283785594 0.0036046152283785594 0.006948905630233565 0.006948905630233565
0.019772105226155662 0.4862 0.0027424439024956275 0.0027424439024956275 0.007069081013913758 0.007069081013913758
};
\addlegendentry{Assume normal}
\addplot+ [
, error bars/.cd, 
x dir=both, x explicit, y dir=both, y explicit]
table [
x error plus=ex+, x error minus=ex-, y error plus=ey+, y error minus=ey-
] {
x y ex+ ex- ey+ ey-
0.05675463383042475 0.9854 0.006381874200695258 0.006381874200695258 0.0016964502906710076 0.0016964502906710076
0.029783665217331123 0.96 0.004328833952023721 0.004328833952023721 0.002771558461815559 0.002771558461815559
0.014713087880870836 0.8962 0.0029827123860535604 0.0029827123860535604 0.004313795749308628 0.004313795749308628
0.009642380002978125 0.7346 0.0023853674802869727 0.0023853674802869727 0.006245027467803034 0.006245027467803034
0.0021140562847594543 0.5224 0.0011203034712754722 0.0011203034712754722 0.007064674792655703 0.007064674792655703
};
\addlegendentry{Omniscient}
\end{axis}

\end{tikzpicture} \\
        \vspace{3mm}
        \begin{tikzpicture}[]
\begin{axis}[height = {6cm},
    title = {Scenario 3: Partially Corellated ($\rho=0.75$)},
    ylabel = {Fraction successful}, ymax = {1.0}, xlabel = {Hard brakes per kilometer}, legend style={at={(0.5,1.05)}, anchor=south, font=\footnotesize}, legend columns=2, grid=both, xticklabel style={/pgf/number format/fixed, /pgf/number format/precision=2}, ymin = {0.0}, width = {7cm}]\addplot+ [
, error bars/.cd, 
x dir=both, x explicit, y dir=both, y explicit]
table [
x error plus=ex+, x error minus=ex-, y error plus=ey+, y error minus=ey-
] {
x y ex+ ex- ey+ ey-
0.0831451030883629 0.9818 0.007114079655369576 0.007114079655369576 0.0018906260585791856 0.0018906260585791856
0.049588695018966185 0.9416 0.005495017124049245 0.005495017124049245 0.0033166380594818385 0.0033166380594818385
0.038041083261445195 0.854 0.004727003229891565 0.004727003229891565 0.004994175443220966 0.004994175443220966
0.016638561050694296 0.6816 0.002693404962471551 0.002693404962471551 0.006588851993641919 0.006588851993641919
0.015922997816150336 0.4798 0.0030510754966022453 0.0030510754966022453 0.007066001533790888 0.007066001533790888
};
\addplot+ [
, error bars/.cd, 
x dir=both, x explicit, y dir=both, y explicit]
table [
x error plus=ex+, x error minus=ex-, y error plus=ey+, y error minus=ey-
] {
x y ex+ ex- ey+ ey-
0.08898230662559689 0.7314 0.007199203137779685 0.007199203137779685 0.006268864947783357 0.006268864947783357
0.0423851053519722 0.5782 0.0048715579593175936 0.0048715579593175936 0.006984748337761956 0.006984748337761956
0.02022593684885273 0.378 0.003194299329247237 0.003194299329247237 0.006858032263069643 0.006858032263069643
0.009668571203540983 0.1726 0.0017731878923584546 0.0017731878923584546 0.005344863095749249 0.005344863095749249
0.007852578628305816 0.0672 0.0014959428905968657 0.0014959428905968657 0.0035410930329452914 0.0035410930329452914
};
\addplot+ [
, error bars/.cd, 
x dir=both, x explicit, y dir=both, y explicit]
table [
x error plus=ex+, x error minus=ex-, y error plus=ey+, y error minus=ey-
] {
x y ex+ ex- ey+ ey-
0.08350450937311038 0.9846 0.007510710816478812 0.007510710816478812 0.001741601169879885 0.001741601169879885
0.05546440644100046 0.9532 0.005803935970743728 0.005803935970743728 0.0029872624168876577 0.0029872624168876577
0.03663720575250644 0.863 0.005072294801460488 0.005072294801460488 0.0048632222020011835 0.0048632222020011835
0.02179862080641367 0.6878 0.003573426729756362 0.003573426729756362 0.006553992902391117 0.006553992902391117
0.01831218661199491 0.4752 0.0037142551685492467 0.0037142551685492467 0.0070630708189765926 0.0070630708189765926
};
\addplot+ [
, error bars/.cd, 
x dir=both, x explicit, y dir=both, y explicit]
table [
x error plus=ex+, x error minus=ex-, y error plus=ey+, y error minus=ey-
] {
x y ex+ ex- ey+ ey-
0.10391648370554485 0.9804 0.008874239757182123 0.008874239757182123 0.0019605960285988197 0.0019605960285988197
0.08175580010229523 0.9516 0.007724653344890724 0.007724653344890724 0.003035346877398558 0.003035346877398558
0.05283994220509099 0.8614 0.005809829961129406 0.005809829961129406 0.004887001591662435 0.004887001591662435
0.03898666963114873 0.691 0.005368692950549745 0.005368692950549745 0.006535468037447208 0.006535468037447208
0.029779406469016625 0.491 0.004362096969349006 0.004362096969349006 0.00707062930436548 0.00707062930436548
};
\addplot+ [
, error bars/.cd, 
x dir=both, x explicit, y dir=both, y explicit]
table [
x error plus=ex+, x error minus=ex-, y error plus=ey+, y error minus=ey-
] {
x y ex+ ex- ey+ ey-
0.1073320377110703 0.8714 0.008637398563332814 0.008637398563332814 0.004734647967724278 0.004734647967724278
0.07699317128339683 0.8182 0.00679509030021899 0.00679509030021899 0.005454878838308729 0.005454878838308729
0.07396993712098031 0.7088 0.00610758714324605 0.00610758714324605 0.00642563380173123 0.00642563380173123
0.0531187846661916 0.5766 0.005255076500404298 0.005255076500404298 0.0069882941588809795 0.0069882941588809795
0.04426741706462751 0.468 0.00432639641808174 0.00432639641808174 0.007057277168441554 0.007057277168441554
};
\addplot+ [
, error bars/.cd, 
x dir=both, x explicit, y dir=both, y explicit]
table [
x error plus=ex+, x error minus=ex-, y error plus=ey+, y error minus=ey-
] {
x y ex+ ex- ey+ ey-
0.05911726368909217 0.986 0.007133118891728767 0.007133118891728767 0.0016617317083254147 0.0016617317083254147
0.03906509256487092 0.96 0.0051688472410064265 0.0051688472410064265 0.0027715584618155584 0.0027715584618155584
0.01668998208126091 0.9022 0.0032762838268948443 0.0032762838268948443 0.004201257206173249 0.004201257206173249
0.016254632652627846 0.7302 0.0033975262242862186 0.0033975262242862186 0.006277696543699677 0.006277696543699677
0.007004856650780219 0.5066 0.002204575783184311 0.002204575783184311 0.007071158904850855 0.007071158904850855
};
\end{axis}

\end{tikzpicture}
    \end{center}
    \vspace{-3mm}
    \caption{Average hard braking frequency and success rate.} \label{fig:bpkm}
\end{figure}

One compelling explanation for the kinks mentioned above is that, as $\lambda$ is increased, since the planner is penalized more severely for unsafe actions, it plans a more conservative trajectory and stays on the road longer.
The longer time on the road gives more chances for unsafe events to occur which are difficult for the planner to avoid because of its inaccurate model.
This explanation is corroborated by the results in \cref{fig:bpkm}.
In both places where there were previously kinks, the number of hard brakes per kilometer decreases as $\lambda$ increases.

%

%
    
\subsection{Correlation Comparison}

It is also interesting to consider the effect that the correlation between model parameters described in \cref{sec:dist} has on the relative effectiveness of the control approaches.
\Cref{fig:gaps}, Scenario 1, shows that when there is no correlation, QMDP offers a significant advantage over mean state MDP, and POMCPOW offers a further significant advantage over QMDP.
In this case, since the parameters are uncorrelated, there is a large amount of uncertainty in them even when some (e.g. $\dot{x}_0$) are easy to observe, and since POMCPOW is able to plan into the future considering this uncertainty, it performs better.
On the other hand, when the parameters are fully correlated as shown in \cref{fig:gaps}, Scenario 2, all of the parameters are easy to estimate by observing only a few, so there is not a significant performance gap between MSM, QMDP, and POMCPOW; all are able to close the gap and achieve nearly the same performance as the upper bound.
\cref{fig:gaps}, Scenario 3, shows the expected behavior between the extremes.

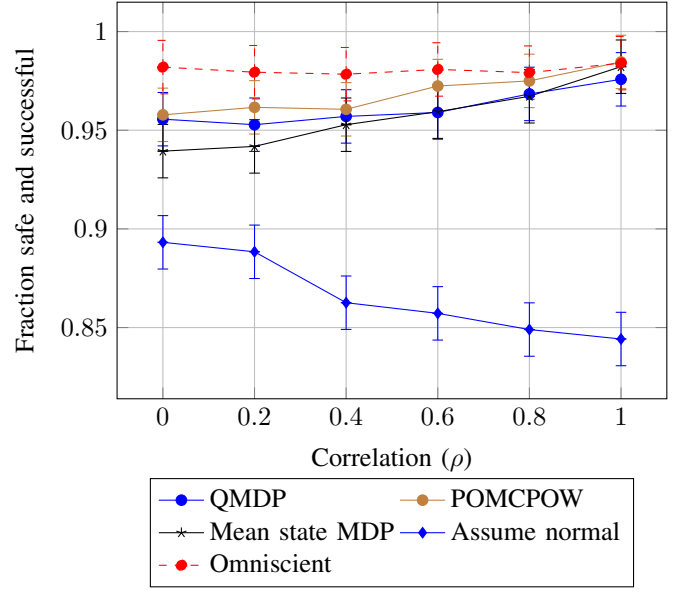
\begin{figure}[tbp]
    \centering
    \begin{tikzpicture}[]
    \begin{axis}[height = {2.7in}, ylabel = {Fraction safe and successful}, xlabel = {Correlation ($\rho$)}, legend style={at={(0.5,-0.2)}, anchor=north}, legend columns=2, grid=both, width = {3.5in}]\addplot+ [
mark = {*}, blue, mark options={fill=blue}, error bars/.cd, 
x dir=both, x explicit, y dir=both, y explicit]
table [
x error plus=ex+, x error minus=ex-, y error plus=ey+, y error minus=ey-
] {
x y ex+ ex- ey+ ey-
0.0 0.9556 0.0 0.0 0.01353728726055671 0.01353728726055671
0.2 0.9528 0.0 0.0 0.01353728726055671 0.01353728726055671
0.4 0.957 0.0 0.0 0.01353728726055671 0.01353728726055671
0.6 0.959 0.0 0.0 0.01353728726055671 0.01353728726055671
0.8 0.9684 0.0 0.0 0.01353728726055671 0.01353728726055671
1.0 0.9758 0.0 0.0 0.01353728726055671 0.01353728726055671
};
\addlegendentry{QMDP}
\addplot+ [
mark = {*}, brown, mark options={fill=brown}, error bars/.cd, 
x dir=both, x explicit, y dir=both, y explicit]
table [
x error plus=ex+, x error minus=ex-, y error plus=ey+, y error minus=ey-
] {
x y ex+ ex- ey+ ey-
0.0 0.9578 0.0 0.0 0.01353728726055671 0.01353728726055671
0.2 0.9616 0.0 0.0 0.01353728726055671 0.01353728726055671
0.4 0.9606 0.0 0.0 0.01353728726055671 0.01353728726055671
0.6 0.9724 0.0 0.0 0.01353728726055671 0.01353728726055671
0.8 0.975 0.0 0.0 0.01353728726055671 0.01353728726055671
1.0 0.9846 0.0 0.0 0.01353728726055671 0.01353728726055671
};
\addlegendentry{POMCPOW}
\addplot+ [
mark = {star}, black, mark options={fill=black}, error bars/.cd, 
x dir=both, x explicit, y dir=both, y explicit]
table [
x error plus=ex+, x error minus=ex-, y error plus=ey+, y error minus=ey-
] {
x y ex+ ex- ey+ ey-
0.0 0.9394 0.0 0.0 0.01353728726055671 0.01353728726055671
0.2 0.9418 0.0 0.0 0.01353728726055671 0.01353728726055671
0.4 0.9528 0.0 0.0 0.01353728726055671 0.01353728726055671
0.6 0.9594 0.0 0.0 0.01353728726055671 0.01353728726055671
0.8 0.9672 0.0 0.0 0.01353728726055671 0.01353728726055671
1.0 0.9822 0.0 0.0 0.01353728726055671 0.01353728726055671
};
\addlegendentry{Mean state MDP}
\addplot+ [
mark = {diamond*}, blue, mark options={fill=blue}, error bars/.cd, 
x dir=both, x explicit, y dir=both, y explicit]
table [
x error plus=ex+, x error minus=ex-, y error plus=ey+, y error minus=ey-
] {
x y ex+ ex- ey+ ey-
0.0 0.8932 0.0 0.0 0.01353728726055671 0.01353728726055671
0.2 0.8884 0.0 0.0 0.01353728726055671 0.01353728726055671
0.4 0.8626 0.0 0.0 0.01353728726055671 0.01353728726055671
0.6 0.8572 0.0 0.0 0.01353728726055671 0.01353728726055671
0.8 0.849 0.0 0.0 0.01353728726055671 0.01353728726055671
1.0 0.8442 0.0 0.0 0.01353728726055671 0.01353728726055671
};
\addlegendentry{Assume normal}
\addplot+ [
mark = {*}, red, dashed, mark options={fill=red, dashed}, error bars/.cd, 
x dir=both, x explicit, y dir=both, y explicit]
table [
x error plus=ex+, x error minus=ex-, y error plus=ey+, y error minus=ey-
] {
x y ex+ ex- ey+ ey-
0.0 0.982 0.0 0.0 0.01353728726055671 0.01353728726055671
0.2 0.9794 0.0 0.0 0.01353728726055671 0.01353728726055671
0.4 0.9784 0.0 0.0 0.01353728726055671 0.01353728726055671
0.6 0.9808 0.0 0.0 0.01353728726055671 0.01353728726055671
0.8 0.9792 0.0 0.0 0.01353728726055671 0.01353728726055671
1.0 0.984 0.0 0.0 0.01353728726055671 0.01353728726055671
};
\addlegendentry{Omniscient}
\end{axis}

\end{tikzpicture}
    \vspace{-0.5cm}
    \caption[Performance variation with $\Beh$ correlation]{Performance variation with $\Beh$ correlation. Error bars indicate the \SI{68}{\percent} (corresponding to one standard deviation in a normal distribution) confidence region determined by the Hoeffding bound. The Naive MDP performance is not shown because it is significantly lower than the other approaches.}
    \label{fig:corplot}
\end{figure}

\Cref{fig:corplot} shows the performance gaps at more points between $\rho=0$ and $1$.
As the correlation increases, the approximate POMDP planning approaches get steadily closer to closing the performance gap with the upper bound.
These results have significant implications for the real world.
It suggests that if most human driver behavior is correlated with easily measurable quantities, near-optimal performance can be achieved by simpler approaches like MSM.
If there is little correlation, more advanced planners that carry the uncertainty further into the future are needed.

\subsection{Robustness}

In the experiments above, internal parameter distribution is assumed to be known exactly.
Because there will be differences between any model used in planning and the way human drivers actually behave, it is important to test robustness.
This section contains tests in which the parameter distribution for planning differs from the true distribution.

\subsubsection{Parameter Correlation Robustness}

\begin{figure}[htpb]
    \centering
    \begin{tikzpicture}[]
    \begin{axis}[height = {2.7in}, ylabel = {Fraction safe and successful}, xlabel = {True correlation ($\rho_\text{sim}$)}, legend style={at={(0.4,-0.2)}, anchor=north}, legend columns=2, grid=both, ymin = {0.9}, width = {3.5in}]\addplot+ [
mark = {square*}, blue, mark options={fill=blue}, error bars/.cd, 
x dir=both, x explicit, y dir=both, y explicit]
table [
x error plus=ex+, x error minus=ex-, y error plus=ey+, y error minus=ey-
] {
x y ex+ ex- ey+ ey-
0.0 0.9544 0.0 0.0 0.01353728726055671 0.01353728726055671
0.2 0.9528 0.0 0.0 0.01353728726055671 0.01353728726055671
0.4 0.9554 0.0 0.0 0.01353728726055671 0.01353728726055671
0.6 0.9582 0.0 0.0 0.01353728726055671 0.01353728726055671
0.8 0.9624 0.0 0.0 0.01353728726055671 0.01353728726055671
1.0 0.9626 0.0 0.0 0.01353728726055671 0.01353728726055671
};
\addlegendentry{POMCPOW ($\rho_\text{plan} = 0.0$)}
\addplot+ [
mark = {*}, blue, mark options={fill=blue}, error bars/.cd, 
x dir=both, x explicit, y dir=both, y explicit]
table [
x error plus=ex+, x error minus=ex-, y error plus=ey+, y error minus=ey-
] {
x y ex+ ex- ey+ ey-
0.0 0.9498 0.0 0.0 0.01353728726055671 0.01353728726055671
0.2 0.953 0.0 0.0 0.01353728726055671 0.01353728726055671
0.4 0.9502 0.0 0.0 0.01353728726055671 0.01353728726055671
0.6 0.955 0.0 0.0 0.01353728726055671 0.01353728726055671
0.8 0.956 0.0 0.0 0.01353728726055671 0.01353728726055671
1.0 0.9582 0.0 0.0 0.01353728726055671 0.01353728726055671
};
\addlegendentry{QMDP ($\rho_\text{plan} = 0.0$)}
\addplot+ [
mark = {square*}, red, mark options={fill=red}, error bars/.cd, 
x dir=both, x explicit, y dir=both, y explicit]
table [
x error plus=ex+, x error minus=ex-, y error plus=ey+, y error minus=ey-
] {
x y ex+ ex- ey+ ey-
0.0 0.9452 0.0 0.0 0.01353728726055671 0.01353728726055671
0.2 0.9508 0.0 0.0 0.01353728726055671 0.01353728726055671
0.4 0.9552 0.0 0.0 0.01353728726055671 0.01353728726055671
0.6 0.9598 0.0 0.0 0.01353728726055671 0.01353728726055671
0.8 0.965 0.0 0.0 0.01353728726055671 0.01353728726055671
1.0 0.9814 0.0 0.0 0.01353728726055671 0.01353728726055671
};
\addlegendentry{POMCPOW ($\rho_\text{plan} = 1.0$)}
\addplot+ [
mark = {*}, red, mark options={fill=red}, error bars/.cd, 
x dir=both, x explicit, y dir=both, y explicit]
table [
x error plus=ex+, x error minus=ex-, y error plus=ey+, y error minus=ey-
] {
x y ex+ ex- ey+ ey-
0.0 0.9434 0.0 0.0 0.01353728726055671 0.01353728726055671
0.2 0.9502 0.0 0.0 0.01353728726055671 0.01353728726055671
0.4 0.9508 0.0 0.0 0.01353728726055671 0.01353728726055671
0.6 0.9582 0.0 0.0 0.01353728726055671 0.01353728726055671
0.8 0.9664 0.0 0.0 0.01353728726055671 0.01353728726055671
1.0 0.9776 0.0 0.0 0.01353728726055671 0.01353728726055671
};
\addlegendentry{QMDP ($\rho_\text{plan} = 1.0$)}
\addplot+ [
mark = {diamond*}, black, dashed, mark options={fill=black, dashed}, error bars/.cd, 
x dir=both, x explicit, y dir=both, y explicit]
table [
x error plus=ex+, x error minus=ex-, y error plus=ey+, y error minus=ey-
] {
x y ex+ ex- ey+ ey-
0.0 0.981 0.0 0.0 0.01353728726055671 0.01353728726055671
0.2 0.977 0.0 0.0 0.01353728726055671 0.01353728726055671
0.4 0.9784 0.0 0.0 0.01353728726055671 0.01353728726055671
0.6 0.98 0.0 0.0 0.01353728726055671 0.01353728726055671
0.8 0.978 0.0 0.0 0.01353728726055671 0.01353728726055671
1.0 0.983 0.0 0.0 0.01353728726055671 0.01353728726055671
};
\addlegendentry{Omniscient}
\end{axis}

\end{tikzpicture}
    \vspace{-0.5cm}
    \caption[Parameter correlation robustness study]{Parameter correlation robustness study. Error bars indicate the \SI{68}{\percent} confidence region determined by the Hoeffding bound.}
    \label{fig:corrob}
\end{figure}
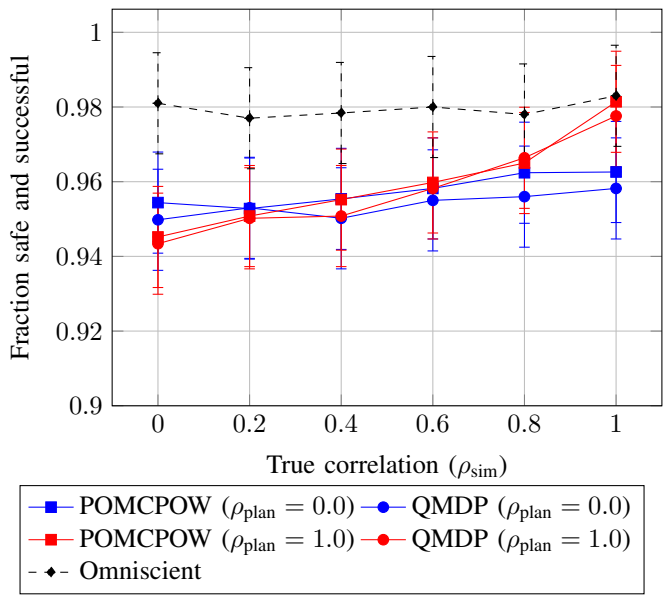

The first robustness test examines the effect of correlation inaccuracy.
POMCPOW and QMDP planners that assume no correlation and full correlation are tested against simulation models with varying levels of correlation.
\Cref{fig:corrob} shows the results.
Performance does not degrade abruptly when the correlation model is inaccurate, though there is an advantage to planning with a correlated model when the true parameters are fully correlated.

\subsubsection{Parameter Domain Robustness}

In the second robustness test, the domain from which the parameters are drawn is expanded by a variable factor.
Specifically, in the new test distribution, the ``Normal'' values from \cref{tab:params} are still used as the midpoints of the distribution, but the distance to the extremes is multiplied by the expansion factor. 
For example, if the expansion factor is \num{2}, the marginal distribution of the desired time gap, $T$, is uniform between \SI{0.5}{\second} and \SI{2.5}{\second}.
Distributions that contain physically nonsensical values such as a negative jam distance, $g_0$, are truncated. 
The planner always uses the distribution defined by values from \cref{tab:params}.

\begin{figure}[htpb]
    \centering
    \begin{tikzpicture}[]
    \begin{axis}[height = {2.7in}, ylabel = {Fraction safe and successful},
        xlabel = {Parameter domain expansion factor\\ (true domain / domain used for planning)},
        xlabel style={align=center},
        legend style={at={(0.5,-0.3)}, anchor=north}, legend columns=2, grid=both, ymin = {0.7}, width = {3.5in}]\addplot+ [
mark = {*}, blue, mark options={fill=blue}, error bars/.cd, 
x dir=both, x explicit, y dir=both, y explicit]
table [
x error plus=ex+, x error minus=ex-, y error plus=ey+, y error minus=ey-
] {
x y ex+ ex- ey+ ey-
0.2 0.9778 0.0 0.0 0.01353728726055671 0.01353728726055671
0.4 0.9768 0.0 0.0 0.01353728726055671 0.01353728726055671
0.6 0.976 0.0 0.0 0.01353728726055671 0.01353728726055671
0.8 0.9754 0.0 0.0 0.01353728726055671 0.01353728726055671
1.0 0.9658 0.0 0.0 0.01353728726055671 0.01353728726055671
1.2 0.955 0.0 0.0 0.01353728726055671 0.01353728726055671
1.4 0.937 0.0 0.0 0.01353728726055671 0.01353728726055671
1.6 0.9058 0.0 0.0 0.01353728726055671 0.01353728726055671
1.8 0.8402 0.0 0.0 0.01353728726055671 0.01353728726055671
2.0 0.7614 0.0 0.0 0.01353728726055671 0.01353728726055671
};
\addlegendentry{QMDP}
\addplot+ [
mark = {square*}, red, mark options={fill=red}, error bars/.cd, 
x dir=both, x explicit, y dir=both, y explicit]
table [
x error plus=ex+, x error minus=ex-, y error plus=ey+, y error minus=ey-
] {
x y ex+ ex- ey+ ey-
0.2 0.8658 0.0 0.0 0.01353728726055671 0.01353728726055671
0.4 0.814 0.0 0.0 0.01353728726055671 0.01353728726055671
0.6 0.747 0.0 0.0 0.01353728726055671 0.01353728726055671
0.8 0.6276 0.0 0.0 0.01353728726055671 0.01353728726055671
1.0 0.515 0.0 0.0 0.01353728726055671 0.01353728726055671
1.2 0.4552 0.0 0.0 0.01353728726055671 0.01353728726055671
1.4 0.39 0.0 0.0 0.01353728726055671 0.01353728726055671
1.6 0.345 0.0 0.0 0.01353728726055671 0.01353728726055671
1.8 0.3112 0.0 0.0 0.01353728726055671 0.01353728726055671
2.0 0.2908 0.0 0.0 0.01353728726055671 0.01353728726055671
};
\addlegendentry{Naive MDP}
\addplot+ [
mark = {*}, brown, mark options={fill=brown}, error bars/.cd, 
x dir=both, x explicit, y dir=both, y explicit]
table [
x error plus=ex+, x error minus=ex-, y error plus=ey+, y error minus=ey-
] {
x y ex+ ex- ey+ ey-
0.2 0.9798 0.0 0.0 0.01353728726055671 0.01353728726055671
0.4 0.9778 0.0 0.0 0.01353728726055671 0.01353728726055671
0.6 0.9842 0.0 0.0 0.01353728726055671 0.01353728726055671
0.8 0.9782 0.0 0.0 0.01353728726055671 0.01353728726055671
1.0 0.9694 0.0 0.0 0.01353728726055671 0.01353728726055671
1.2 0.9646 0.0 0.0 0.01353728726055671 0.01353728726055671
1.4 0.9438 0.0 0.0 0.01353728726055671 0.01353728726055671
1.6 0.9086 0.0 0.0 0.01353728726055671 0.01353728726055671
1.8 0.8574 0.0 0.0 0.01353728726055671 0.01353728726055671
2.0 0.7822 0.0 0.0 0.01353728726055671 0.01353728726055671
};
\addlegendentry{POMCPOW}
\addplot+ [
mark = {star}, black, mark options={fill=black}, error bars/.cd, 
x dir=both, x explicit, y dir=both, y explicit]
table [
x error plus=ex+, x error minus=ex-, y error plus=ey+, y error minus=ey-
] {
x y ex+ ex- ey+ ey-
0.2 0.979 0.0 0.0 0.01353728726055671 0.01353728726055671
0.4 0.9776 0.0 0.0 0.01353728726055671 0.01353728726055671
0.6 0.9776 0.0 0.0 0.01353728726055671 0.01353728726055671
0.8 0.975 0.0 0.0 0.01353728726055671 0.01353728726055671
1.0 0.9624 0.0 0.0 0.01353728726055671 0.01353728726055671
1.2 0.9502 0.0 0.0 0.01353728726055671 0.01353728726055671
1.4 0.9286 0.0 0.0 0.01353728726055671 0.01353728726055671
1.6 0.8944 0.0 0.0 0.01353728726055671 0.01353728726055671
1.8 0.8374 0.0 0.0 0.01353728726055671 0.01353728726055671
2.0 0.7422 0.0 0.0 0.01353728726055671 0.01353728726055671
};
\addlegendentry{Mean state MDP}
\addplot+ [
mark = {diamond*}, blue, mark options={fill=blue}, error bars/.cd, 
x dir=both, x explicit, y dir=both, y explicit]
table [
x error plus=ex+, x error minus=ex-, y error plus=ey+, y error minus=ey-
] {
x y ex+ ex- ey+ ey-
0.2 0.987 0.0 0.0 0.01353728726055671 0.01353728726055671
0.4 0.9722 0.0 0.0 0.01353728726055671 0.01353728726055671
0.6 0.9502 0.0 0.0 0.01353728726055671 0.01353728726055671
0.8 0.9018 0.0 0.0 0.01353728726055671 0.01353728726055671
1.0 0.8428 0.0 0.0 0.01353728726055671 0.01353728726055671
1.2 0.8114 0.0 0.0 0.01353728726055671 0.01353728726055671
1.4 0.7798 0.0 0.0 0.01353728726055671 0.01353728726055671
1.6 0.7344 0.0 0.0 0.01353728726055671 0.01353728726055671
1.8 0.682 0.0 0.0 0.01353728726055671 0.01353728726055671
2.0 0.6102 0.0 0.0 0.01353728726055671 0.01353728726055671
};
\addlegendentry{Assume normal}
\addplot+ [
mark = {*}, red, dashed, mark options={fill=red, dashed}, error bars/.cd, 
x dir=both, x explicit, y dir=both, y explicit]
table [
x error plus=ex+, x error minus=ex-, y error plus=ey+, y error minus=ey-
] {
x y ex+ ex- ey+ ey-
0.2 0.9868 0.0 0.0 0.01353728726055671 0.01353728726055671
0.4 0.9828 0.0 0.0 0.01353728726055671 0.01353728726055671
0.6 0.9828 0.0 0.0 0.01353728726055671 0.01353728726055671
0.8 0.9826 0.0 0.0 0.01353728726055671 0.01353728726055671
1.0 0.9806 0.0 0.0 0.01353728726055671 0.01353728726055671
1.2 0.9698 0.0 0.0 0.01353728726055671 0.01353728726055671
1.4 0.9632 0.0 0.0 0.01353728726055671 0.01353728726055671
1.6 0.951 0.0 0.0 0.01353728726055671 0.01353728726055671
1.8 0.9288 0.0 0.0 0.01353728726055671 0.01353728726055671
2.0 0.8878 0.0 0.0 0.01353728726055671 0.01353728726055671
};
\addlegendentry{Omniscient}
\end{axis}

\end{tikzpicture}
    \vspace{-0.5cm}
    \caption[Parameter domain robustness study]{Parameter domain robustness study. Error bars indicate the \SI{68}{\percent} confidence region determined by the Hoeffding Bound.}
    \label{fig:domrob}
\end{figure}
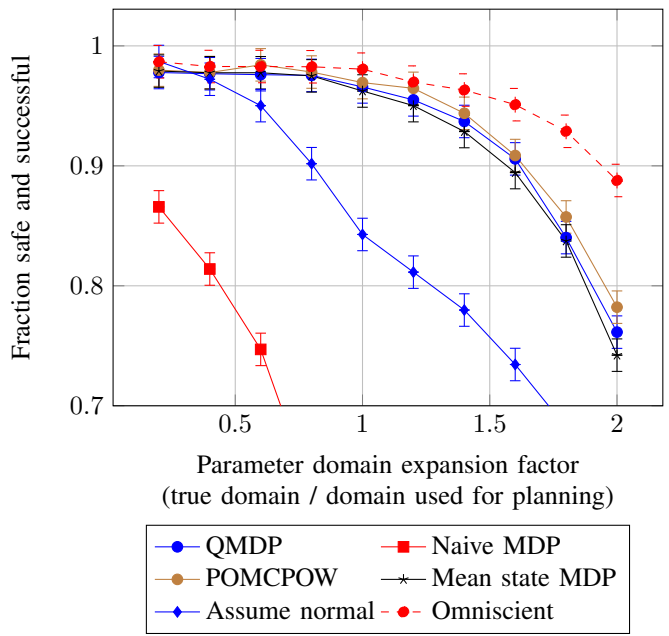

\Cref{fig:domrob} shows the results of tests with expansion factor values between \num{0.2} and \num{2.0}.
All approaches have significantly more success when the expansion factor is less than \num{1}, indicating that the problem is easier in this case, even with an inaccurate model.
There is significant performance degradation at higher expansion factors, both in absolute terms and relative to the omniscient upper bound.
However, this degradation is one-sided; there is little performance lost due to planning with distribution domains that are larger than the true distribution domain.
This one-sidedness suggests that, in practice, prior distributions should be chosen conservatively (i.e. with a larger domain than that of the true distribution) so that the planner can handle situations with both higher- and lower-than-expected levels of uncertainty. 

\section{Conclusion}

This study compared POMDP and MDP planning for autonomous lane changing.
By using the same basic algorithm, MCTS-DPW, for both POMDP and MDP planning, we have isolated the effects of modeling different types of uncertainty.
In particular, the performance of the planners is linked to the degree with which they treat internal state uncertainty as epistemic uncertainty.
The MDP planners ignore epistemic uncertainty and perform the worst.
The POMDP planners are able to consider both aleatoric and epistemic uncertainty and the closer they are able to approximate an optimal POMDP solution, the better they perform.

Though the advantage of the POMDP approaches is clear in all tests, the relative effectiveness of different POMDP approaches depends on the correlation of the distributions of the internal states.
If the internal states are highly correlated, simply estimating them with a particle filter and planning assuming certainty equivalence is adequate to nearly match the upper performance bound.
On the other hand, when the parameters are uncorrelated, the QMDP planner performs much better than the certainty equivalence planner, and POMCPOW performs much better than QMDP.
Moreover, in this uncorrelated case, there is a significant gap between all approaches and the upper bound.

Experiments also characterize the robustness of the algorithms to incorrect parameter distributions.
POMCPOW and QMDP do not suffer significant performance degradation when the parameter correlation is not correct.
Robustness to inaccuracy in the parameter domain is one-sided: when the true domain is larger than that assumed by the planners, performance is adversely affected, but when the true domain is smaller, there is no degradation.

The primary weakness of this investigation is the model of the other drivers.
Since the IDM and MOBIL models were developed to simulate large scale traffic flow~\cite{treiber2000idm,kesting2007mobil}, simulations with these models may not be accurate.
Incorporating models learned from data would further validate the conclusions drawn here.
The model used here also neglects dynamic intentions of other drivers.
Planning based on the possible intentions of other drivers would likely be even more powerful than the approach investigated here because it would enable sophisticated interaction and communication between the autonomous cars and humans.


\section*{Acknowledgments}

Toyota Research Institute (``TRI'')  provided funds to assist the authors with their research but this article solely reflects the opinions and conclusions of its authors and not TRI or any other Toyota entity. The authors would like to thank Christopher Ho for his work in the early stages of this research.


%
%
%

%
\printbibliography


\begin{IEEEbiography}[{\includegraphics[width=1in,height=1.25in,clip,keepaspectratio]{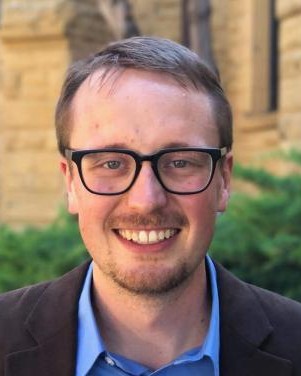}}]{Zachary Sunberg} is an assistant professor in the Ann and H. J. Smead Aerospace Engineering Sciences department at the University of Colorado Boulder. He received Bachelor's and Master's degrees in Aerospace Engineering from Texas A\&M University in 2011 and 2013, and a PhD from Stanford University in 2018. Prior to starting at the University of Colorado, he was a postdoctoral scholar in the department of electrical engineering and computer science at the University of California, Berkeley.
\end{IEEEbiography}

\begin{IEEEbiography}
  [{\includegraphics[width=1in,height=1.25in,clip,keepaspectratio]{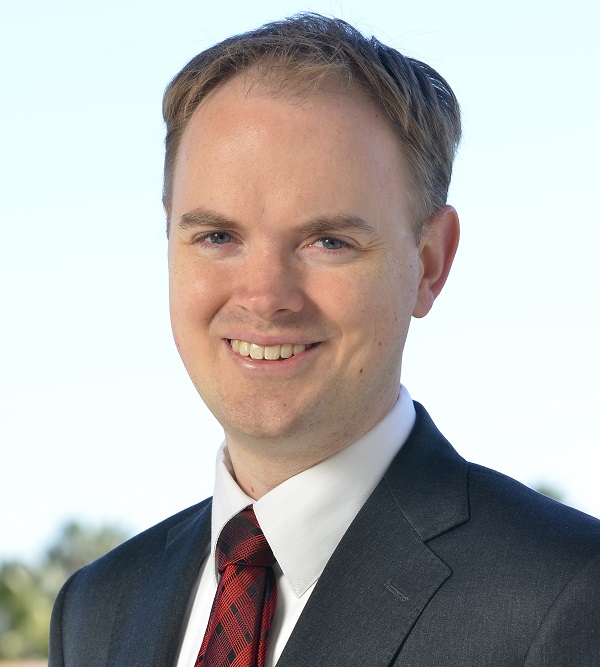}}]{Mykel J. Kochenderfer} received B.S. and M.S. degrees in Computer Science from Stanford University in 2003 and a Ph.D. degree from the University of Edinburgh in 2006. He was a member of the technical staff at MIT Lincoln Laboratory, where he worked on airspace modeling and aircraft collision avoidance. He is now an Associate Professor of Aeronautics and Astronautics at Stanford University and the director of the Stanford Intelligent Systems Laboratory.
\end{IEEEbiography}






\end{document}